\documentclass[runningheads]{llncs}
\usepackage{caption}
\usepackage{subcaption}
\usepackage{amsmath}
\usepackage{array}
\usepackage[T1]{fontenc}
\usepackage{graphicx}
\begin{document}
\title{From Mesh Completion to AI Designed Crown}

\author{Golriz Hosseinimanesh\inst{1}\orcidID{0000-0003-1943-0548} 
\and Farnoosh Ghadiri\inst{2}\orcidID{0000-0002-7232-1888} 
\and Francois Guibault\inst{1}\orcidID{0000-0002-3934-4631} 
\and Farida Cheriet \inst{1}\orcidID{0000-0001-6170-5627} 
\and Julia Keren \inst{3}} 
%
\authorrunning{G. Hosseinimanesh et al.}
%
\institute{Polytechnique Montréal University, Canada \\
\email{\{golriz.hosseinimanesh, francois.guibault, farida.cheriet\}}@polymtl.ca \and
Centre d'intelligence artificielle appliquée (JACOBB), Canada\\
\email{farnoosh.ghadiri@jacobb.ca} \and
Intellident Dentaire Inc., Canada \\
\email{info@kerenor.ca}}

\maketitle             
\begin{abstract} Designing a dental crown is a time-consuming and labor-intensive process. Our goal is to simplify crown design and minimize the tediousness of making manual adjustments while still ensuring the highest level of accuracy and consistency. To this end, we present a new end-to-end deep learning approach, coined Dental Mesh Completion (DMC), to generate a crown mesh conditioned on a point cloud context. The dental context includes the tooth prepared to receive a crown and its surroundings, namely the two adjacent teeth and the three closest teeth in the opposing jaw. We formulate crown generation in terms of completing this point cloud context. A feature extractor first converts the input point cloud into a set of feature vectors that represent local regions in the point cloud. The set of feature vectors is then fed into a transformer to predict a new set of feature vectors for the missing region (crown). Subsequently, a point reconstruction head, followed by a multi-layer perceptron, is used to predict a dense set of points with normals. Finally, a differentiable point-to-mesh layer serves to reconstruct the crown surface mesh. We compare our DMC method to a graph-based convolutional neural network which learns to deform a crown mesh from a generic crown shape to the target geometry. Extensive experiments on our dataset demonstrate the effectiveness of our method, which attains an average of 0.062 Chamfer Distance. 
The code is available at: https://github.com/Golriz-code/DMC.git

\end{abstract}

\section{Introduction} 
\begin{figure}[h]
 \includegraphics[width=\textwidth]{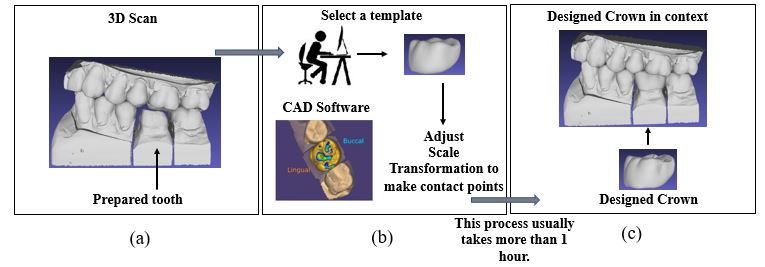}
 \caption{Dental crown design process: a) Dentist prepares the tooth; b) Technician designs the crown; c) Dentist places the crown on the prepared tooth.}
 \label{fig:trad}
\end{figure}

If a tooth is missing, decayed, or fractured, its treatment may require a dental crown. Each crown must be customized to the individual patient in a process, as depicted in Figure \ref{fig:trad}. The manual design of these crowns is a time-consuming and labor-intensive task, even with the aid of computer-assisted design software. Designing natural grooves and ensuring proper contact points with the opposing jaw present significant challenges, often requiring technicians to rely on trial and error. As such, an automated approach capable of accelerating this process and generating crowns with comparable morphology and quality to that of a human expert would be a groundbreaking advancement for the dental industry.

A limited number of studies have focused on how to automate dental crown designs. In \cite{learn1,learn2}, a conditional generative adversarial based network (GAN) is applied to a 2D depth image obtained from a 3D scan to generate a crown for a prepared tooth. Depth images created from a 3D scan can be used directly with 2D convolutional neural networks (CNNs) such as pix2pix \cite{learn3}, which are well-established in computer vision. However, depth images are limited in their ability to capture fine-grained details and can suffer from noise and occlusion issues.
By contrast, point clouds have the advantage of being able to represent arbitrary 3D shapes and can capture fine-grained details such as surface textures and curvatures. \cite{pfnet,min} use point cloud-based networks to generate crowns in the form of 3D point clouds. 
Input point clouds used by \cite{pfnet} are generated by randomly removing a tooth from a given jaw; then the network estimates the missing tooth by utilizing a feature points-based multi-scale generating network. \cite{min} propose a more realistic setting by generating a crown for a prepared tooth instead of a missing tooth. They also incorporate margin line information extracted from the prepared tooth in their network to have a more accurate prediction in the margin line area. The crowns generated by both approaches are represented as point clouds, so another procedure must convert these point clouds into meshes. Creating a high-quality mesh that accurately represents the underlying point cloud data is a challenging task which is not addressed by these two works. 
\cite{Toothcr} proposed a transformer-based network to generate a surface mesh of the crown for a missing tooth. They use two separate networks, one responsible for generating a point cloud and the other for reconstructing a mesh given the crown generated by the first network. 
Similar to \cite{pfnet,min}, the point completion network used by \cite{Toothcr} only uses the Chamfer Distance (CD) loss to learn crown features. This metric's ability to capture shape details in point clouds is limited by the complexity and density of the data. 

Although all aforementioned methods are potentially applicable to the task of dental crown design, most of them fail to generate noise-free point clouds, which is critical for surface reconstruction. One way to alleviate this problem is to directly generate a crown mesh. In \cite{mrnet}, the authors develop a deep learning-based network that directly generates personalized cardiac meshes from sparse contours by iteratively deforming a template mesh, mimicking the traditional 3D shape reconstruction method. To our knowledge, however, the approach in \cite{mrnet} has not been applied to 3D dental scans.  

In this paper, we introduce Dental Mesh Completion (DMC), a novel end-to-end network for directly generating dental crowns without using generic templates. The network employs a transformer-based architecture with self-attention to predict features from a 3D scan of dental preparation and surrounding teeth. These features deform a 2D fixed grid into a 3D point cloud, and normals are computed using a simple MLP. A differentiable point-to-mesh module reconstructs the 3D surface. The process is supervised using an indicator grid function and Chamfer loss from the target crown mesh and point cloud. Extensive experiments validate the effectiveness of our approach, showing superior performance compared to existing methods as measured by the CD metric. 
In summary, our main contributions include proposing the first end-to-end network capable of generating crown meshes for all tooth positions, employing a non-template-based method for mesh deformation (unlike previous works), and showcasing the advantages of using a differentiable point-to-mesh component to achieve high-quality surface meshes.

\section{Related Work}
In the field of 3D computer vision, completing missing regions of point clouds or meshes is a crucial task for many applications. Various methods have been proposed to tackle this problem. Since the introduction of PointNet \cite{pointnet,pointnet++}, numerous methods have been developed for point cloud completion \cite{all}. The recent works PoinTr \cite{pointr} and SnowflakeNet \cite{snowflake} leverage a transformer-based architecture with geometry-aware blocks to generate point clouds. It is hypothesized that using transformers preserves detailed information for point cloud completion. Nonetheless, the predicted point clouds lack connections between points, which complicates the creation of a smooth surface for mesh reconstruction.

Mesh completion methods are usually useful when there are small missing regions or large occlusions in the original mesh data. Common approaches based on geometric priors, self-similarity, or patch encoding can be used to fill small missing regions, as demonstrated in previous studies \cite{small1,small2}, but are not suitable for large occlusions. \cite{meshcompletion} propose a model-based approach that can capture the variability of a particular shape category and enable the completion of large missing regions. However, the resulting meshes cannot achieve the necessary precision required by applications such as dental crown generation.
Having a mesh prior template can also be a solution to generate a complete mesh given a sparse point cloud or a mesh with missing parts. In \cite{mrnet}, cardiac meshes are reconstructed from sparse point clouds using several mesh deformation blocks. Their network can directly generate 3D meshes by deforming a template mesh under the guidance of learned features.

We combine the advantages of point cloud completion techniques with a differentiable surface reconstruction method to generate a dental mesh. Moreover, we used the approach in \cite{mrnet} to directly produce meshes from 3D dental points and compared those results with our proposed method. 
 
\section{Method}
\subsection{Network Architecture}
\begin{figure}[h]
 \includegraphics[width=\textwidth]{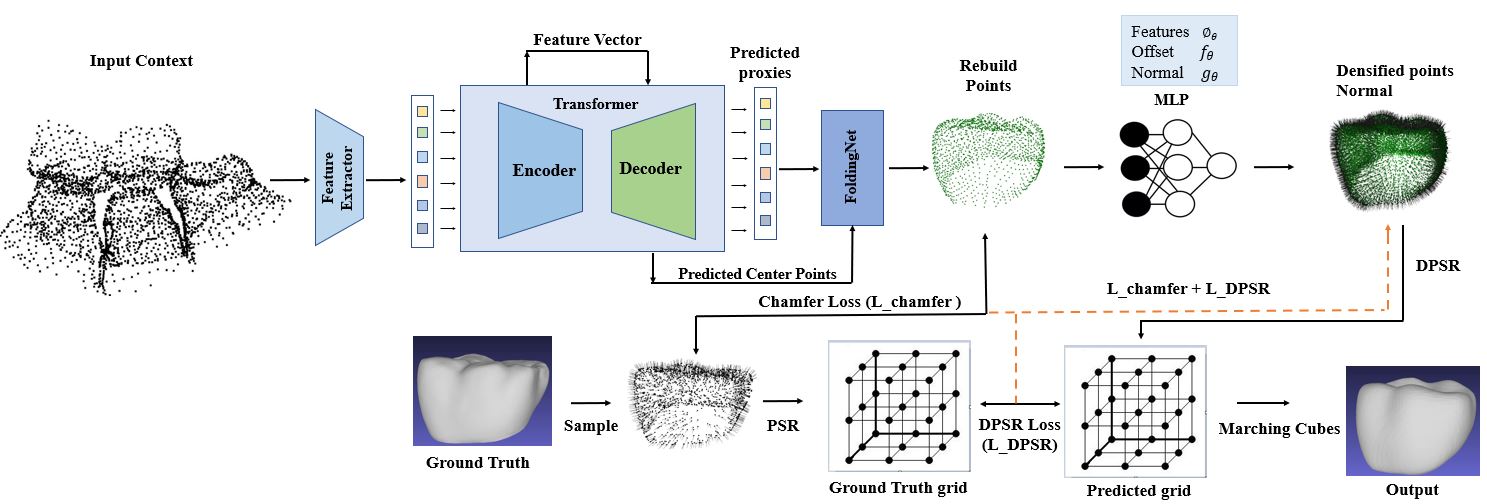}
 \caption{Pipeline of our proposed network.}
 \label{fig:net}
\end{figure}

Our method is an end-to-end supervised framework to generate a crown mesh conditioned on a point cloud context. The overview of our network is illustrated in Figure \ref{fig:net}. The network is characterized by two main components: a transformer encoder-decoder architecture and a mesh completion layer. The following sections explain each part of the network.

\subsubsection{Transformer encoder-decoder}
We adapt the transformer encoder-decoder architecture from \cite{pointr} to extract global and local 3D features from our input (context) using the encoder and generate crown points via the decoder. A dynamic graph convolution network (DGCNN)\cite{DGCNN} is used to group the input points into a smaller set of feature vectors that represent local regions in the context. The generated feature vectors are then fed into the encoder with a geometry-aware block. This block is used to model the local geometric relationships explicitly. The self-attention layer in the encoder updates the feature vectors using both long-range and short-range information. The feature vectors are further updated by a multi-layer perceptron.
The decoder's role is to reason about the crown based on the learnable pairwise interactions between features of the input context and the encoder output. The decoder incorporates a series of transformer layers with a self-attention and cross-attention mechanisms to learn structural knowledge. The output of the transformer decoder is fed into a folding-based decoder \cite{foldingnet} to deform a canonical 2D grid onto the underlying 3D surface of the crown points.

\subsubsection{Mesh completion layer}
In this stage, to directly reconstruct the mesh from the crown points, we use a differentiable Poisson surface reconstruction (DPSR) method introduced by \cite{SAP}. We reconstruct a 3D surface as the zero level set of an indicator function. The latter consists in a regular 3D point grid associated with values indicating whether a point is inside the underlying shape or not.
To compute this function, We first densify the input unoriented crown points. This is done by predicting additional points and normals for each input point by means of an MLP network. After upsampling the point cloud and predicting normals, the network solves a Poisson partial differential equation (PDE) to recover the indicator function from the densified oriented point cloud. We represent the indicator function as a discrete Fourier basis on a dense grid (of resolution $128^3$ ) and solve the Poisson equation (PDE) with the spectral solver method in \cite{SAP}. 

During training, we obtain the estimated indicator grid from the predicted point cloud by using the differentiable Poisson solver. We similarly acquire the ground truth indicator grid
on a dense point cloud sampled from the ground truth mesh, together with the corresponding normals. The entire pipeline is differentiable, which enables the updating of various elements such as point offsets, oriented normals, and network parameters during the training process.
At inference time, we leverage our trained model to predict normals and offsets using Differentiable Poisson Surface Reconstruction (DPSR) \cite{SAP}, solve for the indicator grid, and finally apply the Marching Cubes algorithm \cite{marchingcubes} to extract the final mesh.

\subsubsection{Loss function}
We use the mean Chamfer Distance (CD) \cite{PCN} to constrain point locations. The CD measures the mean squared distance between two point clouds $S_{1}$ and $S_{2}$. Individual distances are measured between each point and its closest point in the other point set, as described in equation (\ref{eq:chamfer}).
In addition, we minimize the $L_{2}$ distance between the predicted indicator function $x$ and a ground truth indicator function $x_{gt}$, each obtained by solving a Poisson PDE \cite{SAP} on a dense set of points and normals. We can express the Mean Square Error (MSE) loss as equation (\ref{eq:MSE}), where $f_{\theta}(X)$ represents a neural network (MLP) with parameters $\theta$ conditioned on the input point cloud $X$, D is the training data distribution, along with indicator functions $x_{i}$ and point samples $X_{i}$ on the surface of shapes. The sum of the CD and MSE losses is used to train the overall network.
 
\begin{equation}
    \text{CD}(S_1, S_2) = \frac{1}{|S_1|} \sum_{x \in S_1} \min_{y \in S_2} |x - y|^2 + \frac{1}{|S_2|} \sum_{y \in S_2} \min_{x \in S_1} |y - x|^2
    \label{eq:chamfer}
\end{equation}

\begin{equation}
    L_{DPSR}(\theta) = {E_{X_i, x_i \sim D}}||Poisson(f_{\theta}(X_i)) - x_i||_2^2
    \label{eq:MSE}
\end{equation}

\section{Experimental results}
\subsection{Dataset and preprocessing}
Our dataset consisted of 388 training, 97 validation, and 71 test cases, which included teeth in various positions in the jaw such as molars, canines, and incisors. 

The first step in the preprocessing was to generate a context from a given 3D scan. To determine the context for a specific prepared tooth, we employed a pre-trained semantic segmentation model \cite{meshsegmentation} to separate the 3D scan into 14 classes representing the tooth positions in each jaw. From the segmentations, we extracted the two adjacent and three opposing teeth of a given prepared tooth, as well as the surrounding gum tissue. To enhance the training data, we conducted data augmentation on the entire dental context, which included the master arch, opposing arch, and shell, treated as a single entity. Data augmentation involved applying 3D translation, scaling, and rotation, thereby increasing the training set by a factor of 10. To enrich our training set, we randomly sampled 10,240 cells from each context to form the input during training. We provide two types of ground truth: mesh and point cloud crowns. To supervise network training using the ground truth meshes, we calculate the gradient from a loss on an intermediate indicator grid. We use the spectral method from \cite{SAP} to compute the indicator grid for our ground truth mesh.

\subsection{Implementation details}
We adapted the architecture of \cite{pointr} for our transformer encoder-decoder module. For mesh reconstruction, we used Differentiable Poisson Surface Reconstruction (DPSR) from \cite{SAP}. All models were implemented in PyTorch with the AdamW optimizer \cite{Adamw}, using a learning rate of 5e-4 and a batch size of 16. Training the model took 400 epochs and 22 hours on an NVIDIA A100 GPU.

\subsection{Evaluation and metrics}
To evaluate the performance of our network and compare it with point cloud-based approaches, we used the Chamfer distance to measure the dissimilarity between the predicted and ground truth points. We employed two versions of CD: $CD_{L_{1}}$ uses the $L_{1}$-norm, while $CD_{L_{2}}$ uses the $L_{2}$-norm to calculate the distance between two sets of points. Additionally, we used the F-score \cite{fscore} with a distance threshold of 0.3, chosen based on the distance between the predicted and ground truth point clouds. We also used the Mean Square Error (MSE) loss \cite{SAP}  to calculate the similarity between the predicted and ground truth indicator grids or meshes.\par
We conducted experiments to compare our approach with two distinct approaches from the literature, as shown in Table \ref{tab1}. The first such approach, PoinTr+margin line \cite{min}, uses the PoinTr \cite{pointr} point completion method as a baseline and introduces margin line information to their network. To compare our work to \cite{min}, we used its official implementation provided by the author. In the second experiment, PoinTr+graph, we modified the work of \cite{mrnet} to generate a dental crown mesh. To this end, we use deformation blocks in \cite{mrnet} to deform a generic template mesh to output a crown mesh under the guidance of the learned features from PoinTr. The deformation module included three Graph Convolutional Networks (GCNs) as in \cite{pixel}.
\begin{table}
\caption{Comparison of proposed method (DMC) with two alternate methods. Evaluation metrics: $CD_{L_{1}}$, $CD_{L_{2}}$ ; MSE on output meshes; F-Score$@0.3$ .}  
\label{tab1}
\centering
\begin{tabular}{|l|l|l|l|l|}
\hline
Method & CD-L1  ($\downarrow$) & CD-L2  ($\downarrow$) & MSE ($\downarrow$) & $F1^{0.3}$ ($\uparrow$) \\ 
\hline
PoinTr + margin line \cite{min} & 0.065 & 0.018 & & 0.54\\
PoinTr + graph  & 1.99 & 1.51 & & 0.08\\
DMC  & {\bfseries 0.0623} & {\bfseries0.011} & 0.0028 & {\bfseries0.70} \\
\hline
\end{tabular}
\end{table}

\begin{figure}[h]
    \centering   
    \begin{subfigure}[b]{0.15\linewidth}
        \centering
        \includegraphics[width=\linewidth]{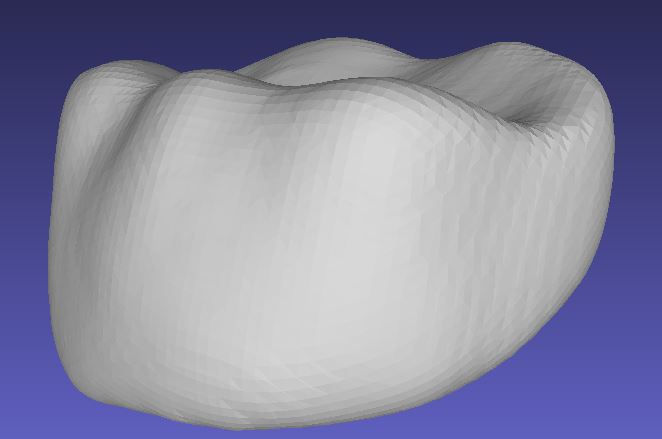}
        \caption{}
    \end{subfigure}
    \hspace{0.05\linewidth}
    \begin{subfigure}[b]{0.15\linewidth}
        \centering
        \includegraphics[width=\linewidth]{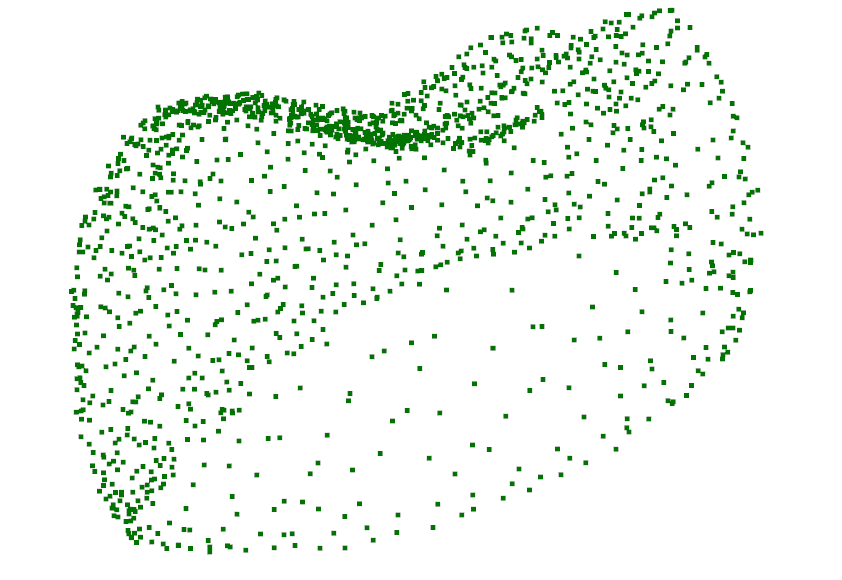}
        \caption{}
    \end{subfigure}
    \caption{Crown mesh predicted by DMC (a) and extracted point set (b). }
    \label{fig:my_label}
\end{figure}

\begin{figure}[h]
    \centering
    
    \begin{subfigure}[b]{0.2\linewidth}
        \centering
        \includegraphics[width=1\linewidth]{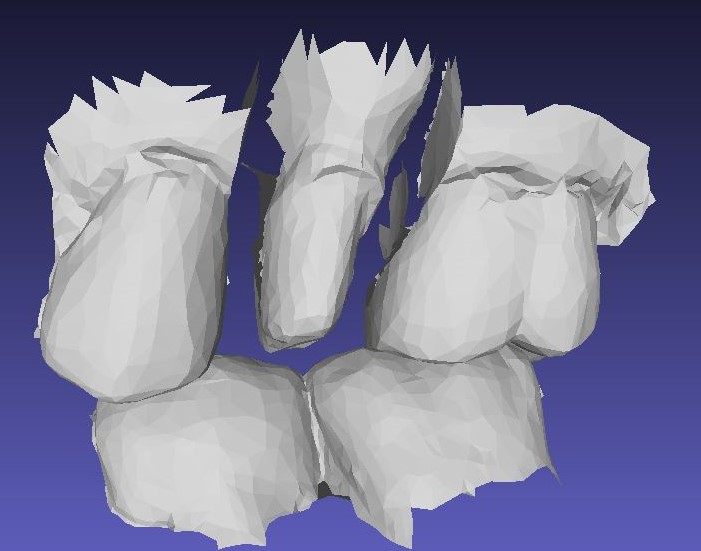}
    \end{subfigure}
    \hspace{0.05\linewidth}
    \begin{subfigure}[b]{0.2\linewidth}
        \centering
        \includegraphics[width=1\linewidth]{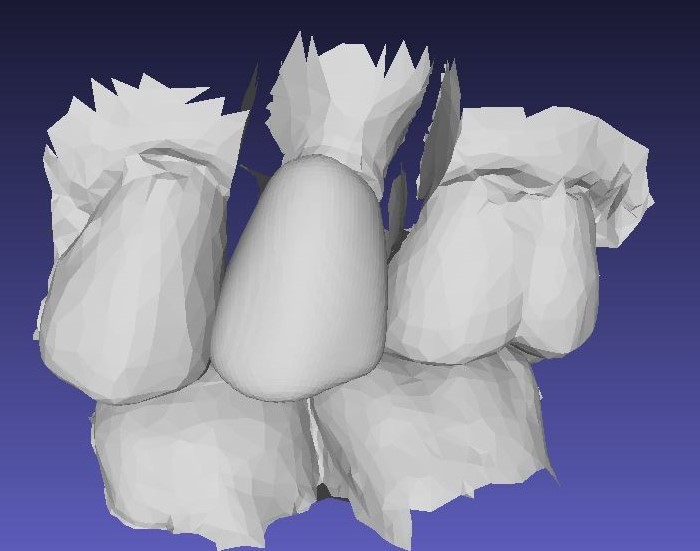}
    \end{subfigure}
    \hspace{0.05\linewidth}
    \begin{subfigure}[b]{0.2\linewidth}
        \centering
        \includegraphics[width=1\linewidth]{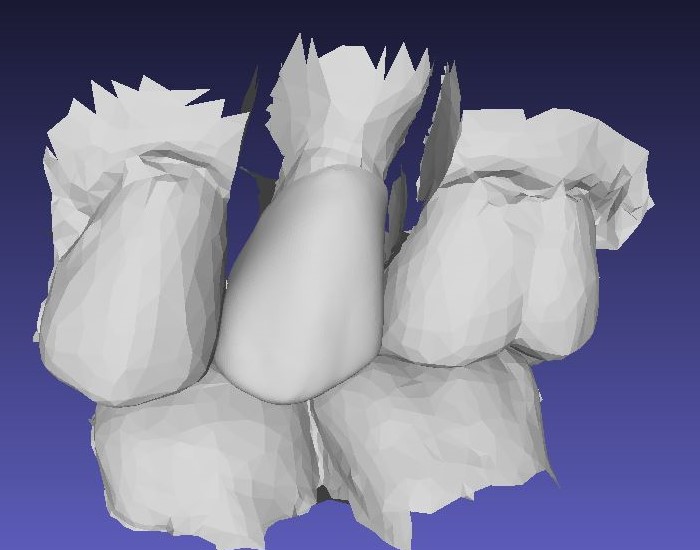}
    \end{subfigure}
    \hspace{0.05\linewidth}
        \begin{subfigure}[b]{0.2\linewidth}
        \centering
        \includegraphics[width=1\linewidth]{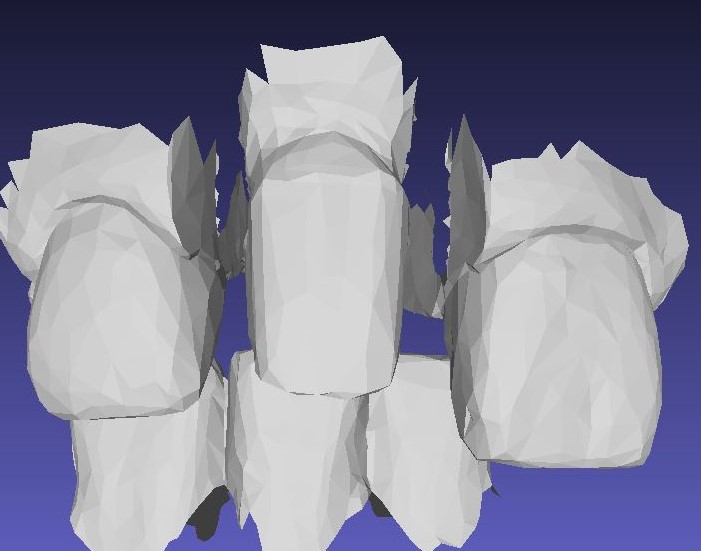}
    \end{subfigure}
    \hspace{0.05\linewidth}
    \begin{subfigure}[b]{0.2\linewidth}
        \centering
        \includegraphics[width=1\linewidth]{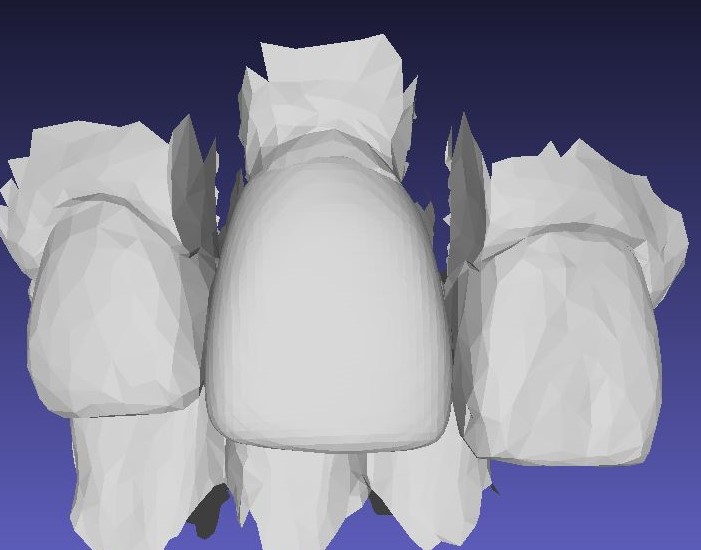}
    \end{subfigure}
    \hspace{0.05\linewidth}
    \begin{subfigure}[b]{0.2\linewidth}
        \centering
        \includegraphics[width=1\linewidth]{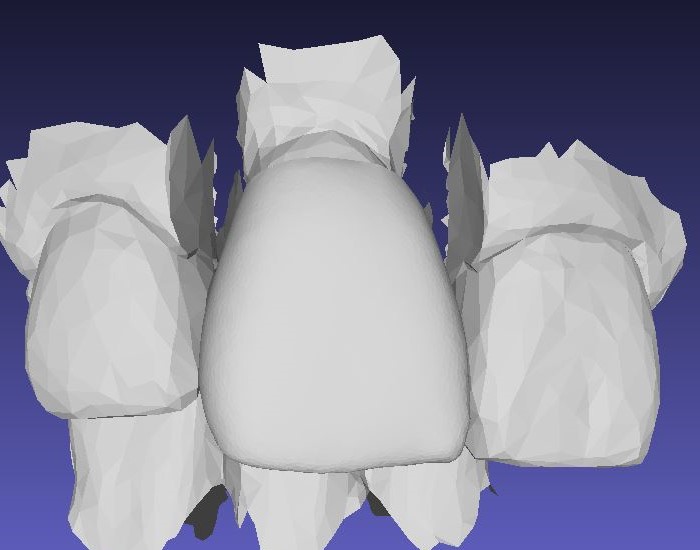}
    \end{subfigure}
    \hspace{0.05\linewidth}
        \begin{subfigure}[b]{0.2\linewidth}
        \centering
        \includegraphics[width=1\linewidth]{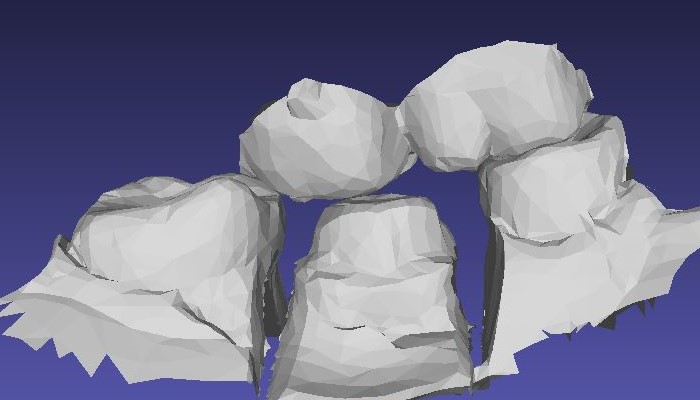}
        \caption{}
    \end{subfigure}
    \hspace{0.05\linewidth}
    \begin{subfigure}[b]{0.2\linewidth}
        \centering
        \includegraphics[width=1\linewidth]{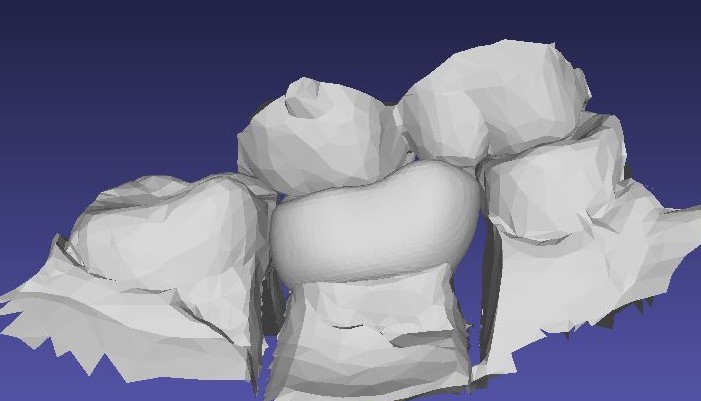}
        \caption{}
    \end{subfigure}
    \hspace{0.05\linewidth}
    \begin{subfigure}[b]{0.2\linewidth}
        \centering
        \includegraphics[width=1\linewidth]{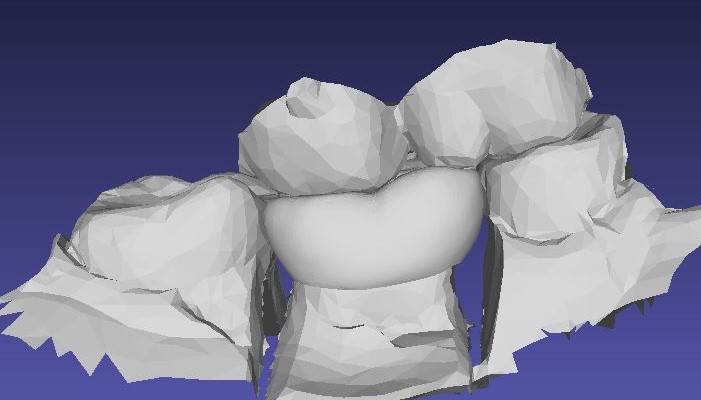}
        \caption{}
    \end{subfigure}
    
    \caption{Examples of mesh completions by the proposed architecture (DMC). a) Input context containing master arch, prepped tooth and opposing arch; b) Generated mesh in its context; c) Ground truth mesh in its context.}
    \label{fig:results}
\end{figure}

\begin{figure}[h]
  \centering
    \hspace{0.05\linewidth}
    \begin{subfigure}[b]{0.15\linewidth}
        \centering
        \includegraphics[width=\linewidth]{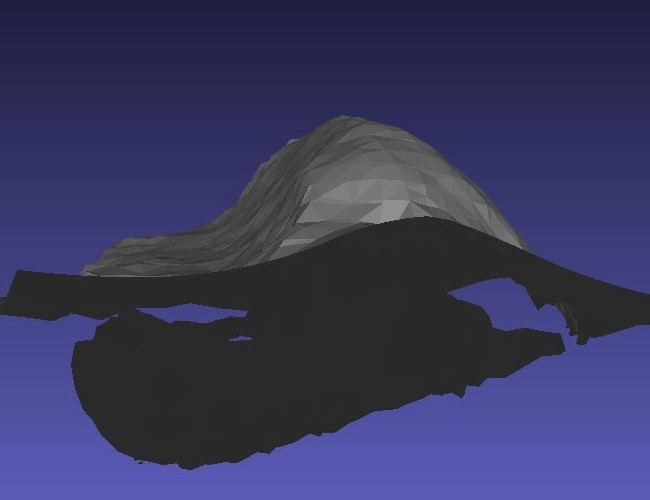}
        \caption{}
    \end{subfigure} 
        \hspace{0.05\linewidth}
     \begin{subfigure}[b]{0.15\linewidth}
        \centering
         \includegraphics[width=\linewidth]{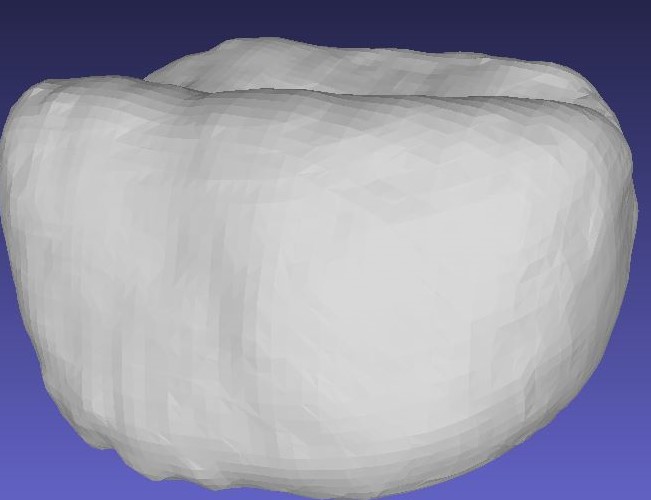}
        \caption{}
    \end{subfigure}
       \hspace{0.05\linewidth}
     \begin{subfigure}[b]{0.15\linewidth}
        \centering
         \includegraphics[width=\linewidth]{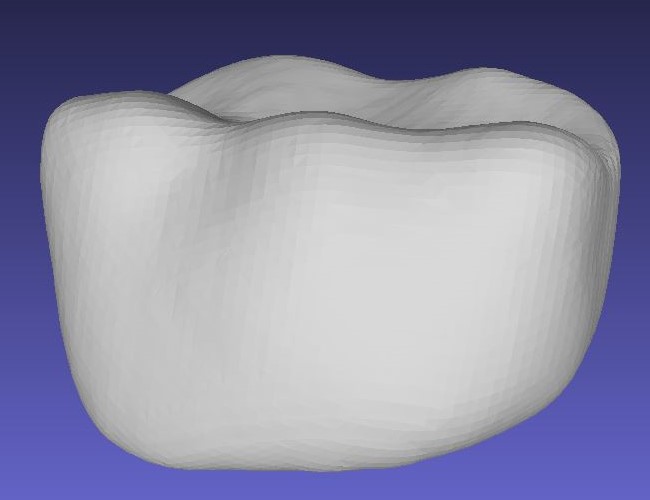}
        \caption{}
    \end{subfigure} 
       \hspace{0.05\linewidth}
     \begin{subfigure}[b]{0.15\linewidth}
        \centering
         \includegraphics[width=\linewidth]{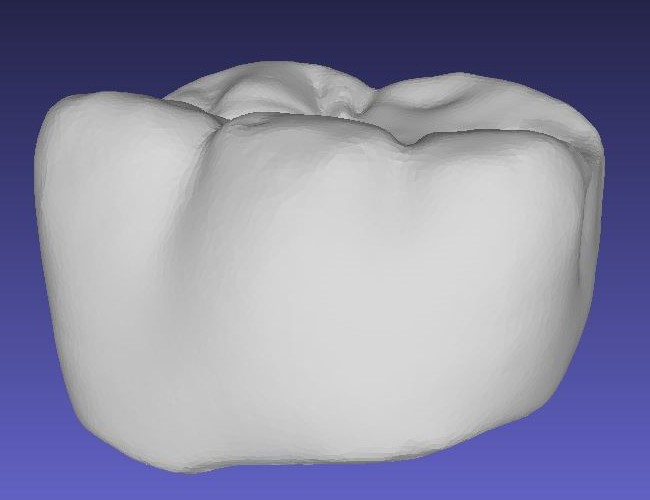}
        \caption{}
    \end{subfigure} 
    \caption{Qualitative comparison of crown mesh generation approaches: a) Standard Poisson surface reconstruction; b) PoinTr \cite{pointr} for point cloud and Shape as points \cite{SAP} for mesh; c) Proposed method (DMC); d) Ground truth shape.}  
    \label{fig:view}
\end{figure}

All experiments used the same dataset, which included all tooth positions, and were trained using the same methodology. To compare the results of the different experiments, we extracted points from the predicted meshes of our proposed network (DMC), as illustrated in Figure \ref{fig:my_label}. Table \ref{tab1} shows that DMC outperforms the two other networks in terms of both CD and F-score. PoinTr+graph achieves poor CD and F-score results compared to the other methods. While the idea of using graph convolutions seems interesting, features extracted from the point cloud completion network don't carry enough information to deform the template into an adequate final crown. Therefore, these methods are highly biased toward the template shape and need extensive pre-processing steps to scale and localize the template.
In the initial two experiments, the MSE metric was not applicable as it was calculated on the output meshes. Figure \ref{fig:results} showcases the visual results obtained from our proposed network (DMC). Furthermore, Figure \ref{fig:view} presents a visual comparison of mesh surfaces generated by various methods for a sample molar tooth.

\subsection{Ablation study}
We conducted an ablation study to evaluate the components of our architecture. We started with the baseline PoinTr \cite{min}, a point completion method. To enhance it, we integrated Shape as Points (SAP) \cite{SAP} as a separate network for mesh reconstruction from the PoinTr-generated point cloud. Next, we tested our proposed method (DMC) by excluding the Mean Square Error (MSE) loss function. Finally, we assessed DMC's performance, including the MSE loss function. The results, shown in Table \ref{tab2}, demonstrate the consistent improvements achieved by our full model across all evaluation metrics.
\begin{table}
\caption{Results of ablation study. Metrics are the same as in Table \ref{tab1}.}  
\label{tab2}
\centering
\begin{tabular}{|l|l|l|l|l|}
\hline
Method & CD-L1  ($\downarrow$) & CD-L2  ($\downarrow$) & MSE ($\downarrow$) & $F1^{0.3}$ ($\uparrow$) \\
\hline
PoinTr \cite{pointr} & 0.070 & 0.023  & & 0.24 \\
PoinTr + SAP  & 0.067 &0.021& 0.031& 0.50\\
DMC without MSE  & 0.0641 & 0.015 & & 0.65\\
DMC (Full Model)& {\bfseries 0.0623} & {\bfseries0.011} & {\bfseries0.0028} & {\bfseries0.70} \\
\hline
\end{tabular}
\end{table}
\section{Conclusion}
Existing deep learning-based dental crown design solutions require additional steps to reconstruct a surface mesh from the generated point cloud. In this study, we propose a new end-to-end approach that directly generates high-quality crown meshes for all tooth positions. By utilizing transformers and a differentiable Poisson surface reconstruction solver, we effectively reason about the crown points and convert them into mesh surfaces using Marching Cubes. Our experimental results demonstrate that our approach produces accurately fitting crown meshes with superior performance. In the future, incorporating statistical features into our deep learning method for chewing functionality, such as surface contacts with adjacent and opposing teeth, could be an interesting avenue to explore.
\section{Acknowledgments}
This work was funded by Kerenor Dental Studio, Intellident Dentaire Inc..

 \bibliographystyle{splncs04}
 \bibliography{mybibliography}
 
\end{document}